\documentclass[runningheads]{llncs}

\usepackage{eccv}

\usepackage{eccvabbrv}
\usepackage{graphicx}
\usepackage{booktabs}
\usepackage{soul}
\usepackage{caption}                
\usepackage{array}                  
\usepackage[table,x11names]{xcolor} 
\usepackage{multirow}               
\usepackage{calc} 
\usepackage{amsmath}
\usepackage{amssymb}

\usepackage[accsupp]{axessibility}  

\usepackage{hyperref}

\usepackage{orcidlink}

\usepackage[most]{tcolorbox}

\usepackage{algorithm}
\usepackage{algorithmic}

\usepackage{subcaption}     
\usepackage{float} 
\usepackage{enumitem}

\usepackage{adjustbox}   
\usepackage{makecell}    
\usepackage{tabularx}    
\usepackage{ragged2e}    

\begin{document}

\title{AViTS: Adaptive Spatiotemporal Token Selection for Efficient Dynamic-Resolution Generation} 

\titlerunning{AViTS: Adaptive Spatiotemporal Token Selection}

\author{Haoran Qin\inst{1,3}$^{\star}$ \and
Zhengan Yan\inst{1}$^{\star}$ \and
Shikang Zheng\inst{1}$^{\star}$ \and
Xiaobing Tu\inst{2} \and
Jiacheng Liu\inst{1} \and
Yuqi Lin\inst{1,4} \and
Chang Zou\inst{1} \and
JinShan Liu\inst{5} \and
Peiliang Cai\inst{1} \and
Xiantao Zhang\inst{2} \and
Jinkui Ren\inst{2} \and
Linfeng Zhang\inst{1}$^{\dagger}$}

\authorrunning{H.~Qin et al.}

\institute{Shanghai Jiao Tong University, China \and
Terminal Intelligent Computing Division, Alibaba Cloud, China \and
Shandong University, China \and
Jilin University, China \and
Xi'an Jiaotong University, China}

\maketitle

\begingroup
\renewcommand{\thefootnote}{}
\footnotetext{$^{\star}$\,Equal contribution.\quad$^{\dagger}$\,Corresponding author.}
\endgroup

\begin{figure}[h]
    \centering
    \vspace{-8mm}
    \includegraphics[width=0.85\linewidth]{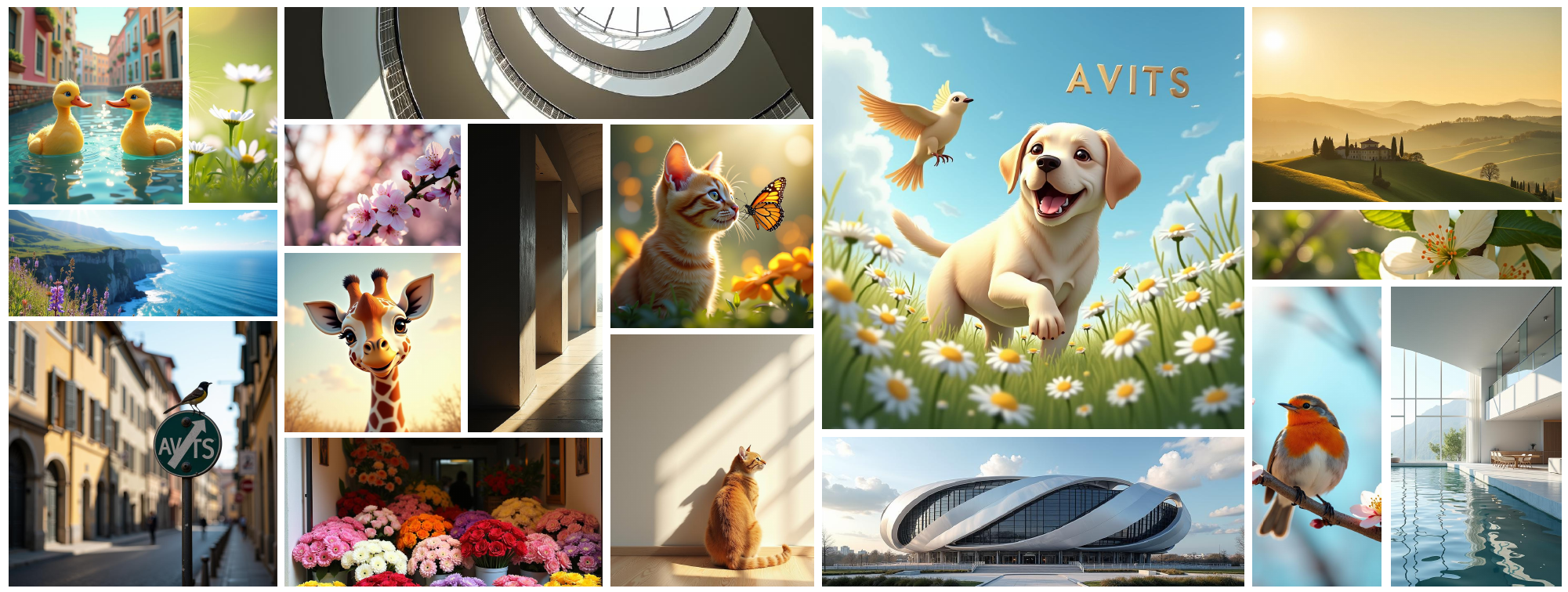}
    \caption{Images sampled by FLUX.1-dev with AViTS with 6.34× acceleration.}
    \label{fig:head}
    \vspace{-13mm}
\end{figure}

\begin{abstract}
  Diffusion Transformers (DiTs) achieve high-quality generation but are costly due to iterative sampling. Dynamic-resolution sampling reduces early-stage cost by denoising at low resolution; however, uniformly upsampling all latent tokens at resolution transitions incurs redundant computation and may degrade fine-detail consistency. Existing partial upsampling strategies typically rely on local latent structure cues or single-step statistics, making it difficult to jointly capture token–text semantic relevance and token-wise representation dynamics across diffusion steps. We propose \textbf{AViTS}, an adaptive spatiotemporal token selection framework for dynamic-\allowbreak resolution DiTs. AViTS models \textbf{spatial importance} via latent–text attention and \textbf{temporal importance} via token-level feature variation across diffusion timesteps, and fuses them to enable \textbf{spatiotemporal importance-aware selective upsampling}: it prioritizes resolution refinement for critical tokens while deferring less important ones, thereby reducing redundant high-resolution computation and improving the quality–efficiency trade-off. AViTS achieves up to \textbf{6.34$\times$} on \textbf{FLUX} and nearly \textbf{9$\times$} FLOPs reduction on \textbf{Qwen-Image-Edit} and \textbf{FLUX.1-Kontext-dev}, orthogonal to distillation, quantization, and feature caching, and reaching \textbf{14.76$\times$} with distilled models. Code: \url{https://github.com/QHR69/AViTS}.
\end{abstract}
\section{Introduction}
\label{sec:intro}

Diffusion models have become a dominant paradigm for image and video generation and editing. Among them, Diffusion Transformers (DiTs) stand out for their scalability and strong performance in high-fidelity conditional generation and editing. However, DiT inference remains costly due to iterative sampling: generating or editing a single sample typically requires dozens of denoising steps, each passing through a large Transformer backbone, resulting in high latency. This challenge is further amplified at high resolutions, where the increased token count leads to quadratic growth in self-attention cost and memory footprint, hindering real-time deployment and use in resource-constrained settings.

\begin{figure}[h]
    \centering
    \includegraphics[width=1.0\linewidth]
    {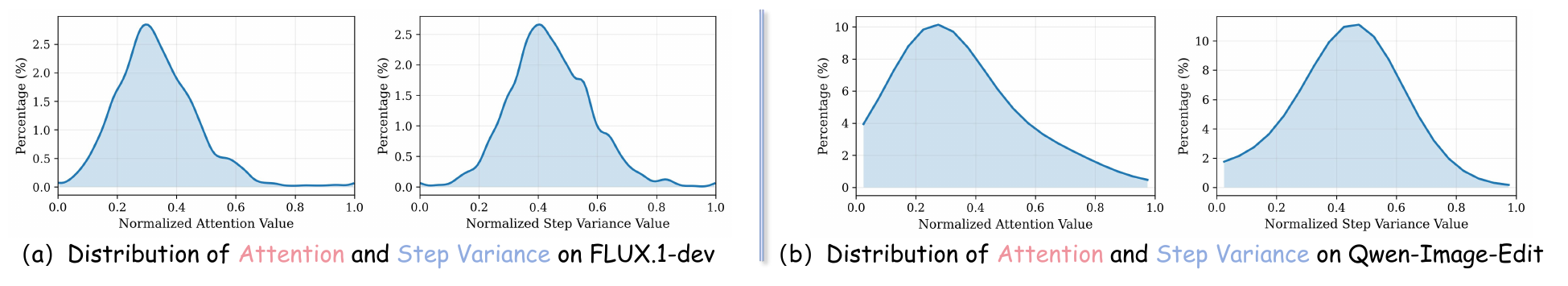}
    \captionof{figure}{Token-importance distributions. (a) Distributions of attention and step variance on FLUX.1-dev. (b) Distributions of attention and step variance on Qwen-Image-Edit.}
    \label{fig:intro}
    \vspace{-12pt}
\end{figure}


To accelerate inference, existing approaches mainly follow two directions including reducing the number of sampling steps and reducing the computational cost of each step. Methods in the first category reduce the number of sampling steps through techniques such as distillation and consistency training. Methods in the second category reduce the per step computational cost through techniques such as sparse computation and feature caching~\cite{toca,duca}. Recently, dynamic resolution sampling has emerged as a promising direction for spatial acceleration in DiTs. It performs denoising at a reduced resolution during early stages to save computation and then gradually transitions to higher resolution to recover fine details. Nevertheless, during resolution transitions many dynamic resolution strategies still upsample all latent tokens and perform high resolution computation uniformly. This process introduces substantial redundant computation and may also cause artifacts and inconsistency across stages. The recent development of partial upsampling further raises an important question. Under a limited high resolution computation budget how should latent tokens be prioritized so that computation focuses on the most critical content and achieves a better balance between efficiency and quality. However, existing partial upsampling methods usually determine priorities using internal latent cues such as spatial structure or local single step statistics through heuristic rules. These approaches do not explicitly model text conditioned semantic alignment or cross timestep representation dynamics~\cite{jeong2025upsample,zheng2026sketch}. As a result they may fail to reliably identify and prioritize the truly critical tokens.

\begin{figure}[htbp]
    \centering
    \includegraphics[width=1.0\linewidth]
    {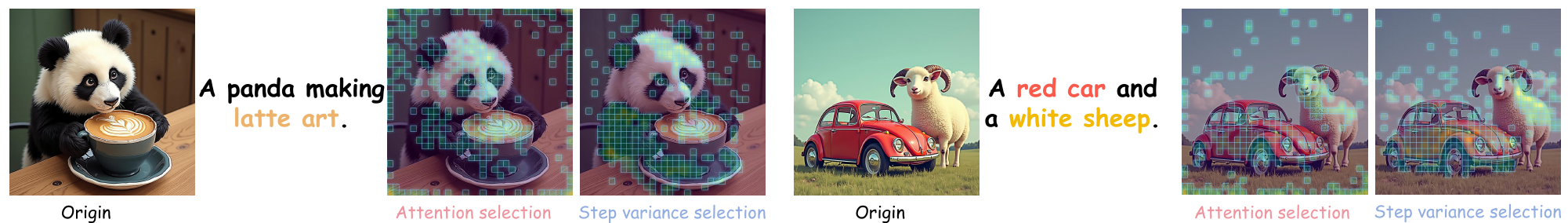}
    \captionof{figure}{Heatmaps of token importance on two text-to-image examples. Attention-based selection covers instruction-relevant semantic regions, whereas step-variance-based selection focuses on regions requiring fine-grained synthesis.}
    \label{fig:intro_add}
    \vspace{-12pt}
\end{figure}

To characterize where token importance arises in low-resolution denoising, we conduct statistical analysis and visualization from two perspectives: semantic alignment and cross-step dynamics. Specifically, we quantify semantic alignment by aggregating token–text cross-attention responses at low resolution, and quantify cross-step dynamics by measuring the magnitude of token representation changes across multiple sampling steps (step variance) in the DiT backbone. As shown in Fig.~\ref{fig:intro}, both the attention-based spatial importance and the step-variance-based temporal importance exhibit pronounced long-tailed, highly non-uniform distributions: only a small fraction of tokens strongly correlate with the text condition, and only a small subset continue to evolve during denoising and remain more sensitive to subsequent updates. This pattern holds consistently on FLUX.1-dev (Fig.~\ref{fig:intro}(a)) and Qwen-Image-Edit (Fig.~\ref{fig:intro}(b)). The corresponding heatmaps further qualitatively suggest that attention tends to cover instruction-relevant semantic regions, whereas higher step variance often concentrates on regions requiring fine-grained modeling or modification (Fig.~\ref{fig:intro_add}). These observations indicate that jointly modeling semantic alignment and cross-step dynamics enables more effective allocation of a limited high-resolution budget.

Motivated by this insight, we propose AViTS (Adaptive Spatiotemporal Token Selection), a spatiotemporal information-driven token selection framework for efficient dynamic-resolution generation. AViTS models token importance along two complementary dimensions: it estimates spatial importance via latent–text attention interactions to capture semantic alignment strength, and estimates temporal importance via token-wise feature variation across diffusion timesteps (step variance) to capture evolution and stability. AViTS then fuses these signals into a unified spatiotemporal importance score and performs importance-aware selective upsampling: it upsamples high-importance tokens earlier while deferring low-importance tokens, reducing redundant high-resolution computation while preserving critical semantics and details. Extensive experiments demonstrate that AViTS achieves a strong efficiency–quality trade-off across generation and editing (Fig.~\ref{fig:head}): it provides up to 6.34× acceleration on FLUX, nearly 9× FLOPs reduction on Qwen-Image-Edit and FLUX.1-Kontext-dev, and up to 14.76× when combined with distillation, while remaining compatible with quantization and feature caching.
Our contributions are three-fold:

\begin{itemize}[leftmargin=10pt,topsep=-2pt,label=\textbf{•}]

    \item \textbf{Spatiotemporal Heterogeneity Analysis.}
    From the perspectives of spatial semantic alignment and temporal cross-step dynamics, we reveal and validate pronounced spatiotemporal heterogeneity of latent-token importance during low-resolution denoising, providing empirical grounding for prioritization in resolution transitions.

    \item \textbf{Spatiotemporal Importance Modeling.}
    We introduce \textbf{AViTS}, formulating upsampling prioritization as a spatiotemporal importance estimation problem jointly determined by semantic alignment and cross-step dynamics, and enabling spatiotemporal-importance-aware selective upsampling.

    \item \textbf{Broad Effectiveness And Composable Acceleration.}
    AViTS substantially reduces inference cost while preserving quality across multiple models and tasks, and composes effectively with distillation, quantization, and feature caching for further acceleration.

\end{itemize}
\section{Related Works}

Diffusion models have become a dominant paradigm for high-quality visual generation~\cite{sohl2015deep}. 
While early methods were built on U-Net backbones~\cite{ronneberger2015u}, recent systems increasingly adopt Diffusion Transformers (DiTs) for improved scalability~\cite{peebles2023scalable}. 
However, the iterative denoising process incurs substantial inference cost, motivating extensive research on diffusion acceleration.

\subsection{Model Compression–Based Acceleration}
Model compression reduces inference cost by simplifying the network structure or numerical representation. 
Representative approaches include structural pruning~\cite{structural_pruning_diffusion}, numerical quantization~\cite{li2023q,shang2023post}, knowledge distillation, and token reduction strategies~\cite{bolya2023token}. 
These techniques lower computation and memory usage with minimal modification to the inference pipeline. 
However, maintaining generation quality usually requires additional fine-tuning, and aggressive compression may degrade generalization under complex editing scenarios.

\subsection{Temporal Acceleration}
Another major direction reduces the number of denoising steps. 
Deterministic samplers such as DDIM~\cite{song2020denoising} and higher-order solvers (e.g., DPM-Solver) improve the efficiency–quality trade-off by controlling numerical errors. 
Alternative formulations, including Rectified Flow, progressive distillation, and consistency models, further shorten denoising trajectories. 
Complementary training-free approaches exploit temporal redundancy by caching or forecasting intermediate features across timesteps~\cite{Liu2025SpeCa,cai2026lesa,liu2025survey}. 
Although effective, many existing caching methods rely on heuristic temporal assumptions and lack principled modeling of token-level dynamics.

\subsection{Spatial Acceleration}
Spatial acceleration reduces computation by lowering latent resolution, which in DiTs corresponds to decreasing the number of spatial tokens. 
Sparse attention patterns~\cite{child2019generating} provide limited gains, while coarse-to-fine strategies perform early denoising at reduced resolution. 
Cascaded diffusion frameworks~\cite{ho2022cascaded} follow this paradigm but require retraining. 
Recent training-free methods selectively upsample tokens during sampling. 
For instance, RALU~\cite{jeong2025upsample} uses edge detection to identify important tokens, while Fresco~\cite{zheng2026sketch} selects tokens based on inter-channel variance. 
However, these low-level heuristics are weakly aligned with editing semantics, often leading to unstable token selection and inconsistent denoising trajectories. 
Bottleneck Sampling~\cite{tian2025training} further adjusts resolution during inference but may introduce trajectory distortion and artifacts after upsampling.

\section{Method}
\label{sec:method}

\subsection{Preliminaries}
\label{sec:prelim}

\paragraph{Setup and token representation.}
We work with Diffusion Transformers (DiTs) that operate in latent space.
An image of resolution $H{\times}W$ is encoded by a VAE into a compact spatial map and packed into a sequence of $M$ non-overlapping patch tokens.
The image token sequence is $\mathbf{Z} = \{\mathbf{z}_i\}_{i=1}^{M} \in \mathbb{R}^{M \times D}$, where $\mathbf{z}_i \in \mathbb{R}^{D}$ is the feature of the $i$-th patch at spatial coordinate $\mathbf{p}_i = (r_i, c_i)$.
The text prompt is encoded into $L$ tokens $\mathbf{C} = \{\mathbf{c}_j\}_{j=1}^{L} \in \mathbb{R}^{L \times D}$.
In models with joint image--text attention, $\mathbf{Z}$ and $\mathbf{C}$ interact in a shared attention mechanism, producing cross-modal attention that we use for spatial importance in \cref{sec:spatial}.

\paragraph{Flow matching.}
We adopt the flow-matching formulation.
A velocity network $v_\theta(\mathbf{x}_t,\,t,\,\mathbf{c})$ is trained to regress the conditional velocity.
At inference, we integrate the learned ODE from $t{=}1$ to $t{=}0$ via Euler steps:
\begin{equation}
  \mathbf{x}_{t_{i+1}} \;=\; \mathbf{x}_{t_i}
  + v_\theta\!\left(\mathbf{x}_{t_i},\,t_i,\,\mathbf{c}\right)\Delta t_i,
  \qquad \Delta t_i = t_{i+1}-t_i < 0.
  \label{eq:euler}
\end{equation}
This update is used at each denoising step in our pipeline.

\begin{figure}[htbp]
    \centering
    \includegraphics[width=1\linewidth]{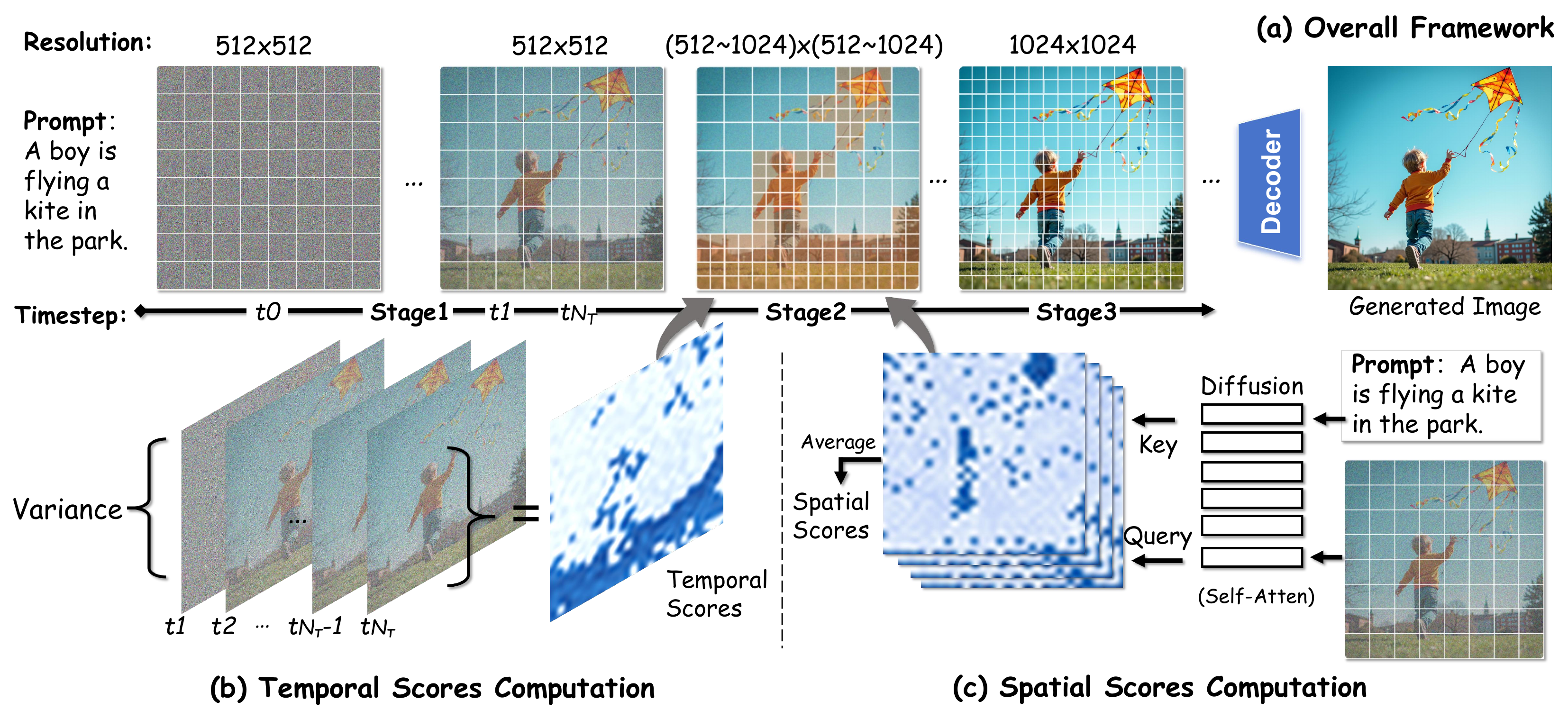}
     \caption{\textbf{Overview of AViTS and spatiotemporal importance estimation.}
\textbf{(a)} Three-stage dynamic-resolution sampling: Stage~1 low-res denoising with signal collection over the last $N_T$ steps, Stage~2 selective upsampling of top-$K$ tokens, and Stage~3 full-res refinement.
\textbf{(b)} Temporal importance from token-wise step variance across the collected snapshots.
\textbf{(c)} Spatial importance from aggregated latent--text cross-attention (averaged over heads/blocks and collection steps); fused scores guide prioritization.}
    \label{fig:avits_framwork}
\end{figure}

\subsection{Framework Overview}
\label{sec:framework}

AViTS organizes inference into three stages of progressively increasing spatial resolution (Figure~\ref{fig:avits_framwork}(a)).
Stage~1 performs low-resolution denoising and includes a feature-collection phase in its latter steps.
This design differs from prior dynamic-resolution methods that lack an explicit multi-step, multimodal collection phase.

\paragraph{Stage~1: Low-resolution denoising and feature collection.}
We initialize with noise $\mathbf{Z}_{t_0}\!\sim\!\mathcal{N}(\mathbf{0},\mathbf{I})$ at the target resolution $H{\times}W$, spatially downsample by factor $f{=}2$, and reduce the token count to $M' = M/f^{2}$.
The velocity network $v_\theta$ performs $N_1$ Euler denoising steps via Eq.~\eqref{eq:euler} to establish coarse global structure at low computational cost.
Over the subsequent $N_T$ denoising steps at the same reduced resolution, we record two complementary signals: the per-step latent snapshots $\{\mathbf{Z}^{(t_k)}\}_{k=1}^{N_T}$, which capture the temporal evolution of each token, and the cross-modal attention maps at each step, which encode the alignment between image tokens and the text condition.
These signals feed the spatiotemporal importance estimator detailed in~\cref{sec:importance}.

\paragraph{Stage~2: Selective upsampling.}
Using the collected signals, AViTS computes a scalar importance score $S_i$ for each of the $M'$ tokens and selects the top-$K$ subset $\mathcal{S}$ with $K = \lfloor\rho M'\rfloor$, $\rho \in (0,1)$.
Tokens in $\mathcal{S}$ are expanded via orthogonal upsampling; tokens in $\bar{\mathcal{S}}$ remain at the current resolution.
We re-inject coordinate-bound noise into the mixed-resolution sequence and perform $N_2$ denoising steps.

\paragraph{Stage~3: Full-resolution refinement.}
The remaining tokens $\bar{\mathcal{S}}$ are expanded to complete the $M$-token sequence.
After a final coordinate-bound noise re-injection and reordering by spatial coordinate, $N_3$ denoising steps recover fine details.

\subsection{Spatiotemporal Token Importance Estimation}
\label{sec:importance}

\begin{figure}[h]
    \centering
    \vspace{-8mm}
    \includegraphics[width=1.0\linewidth]
    {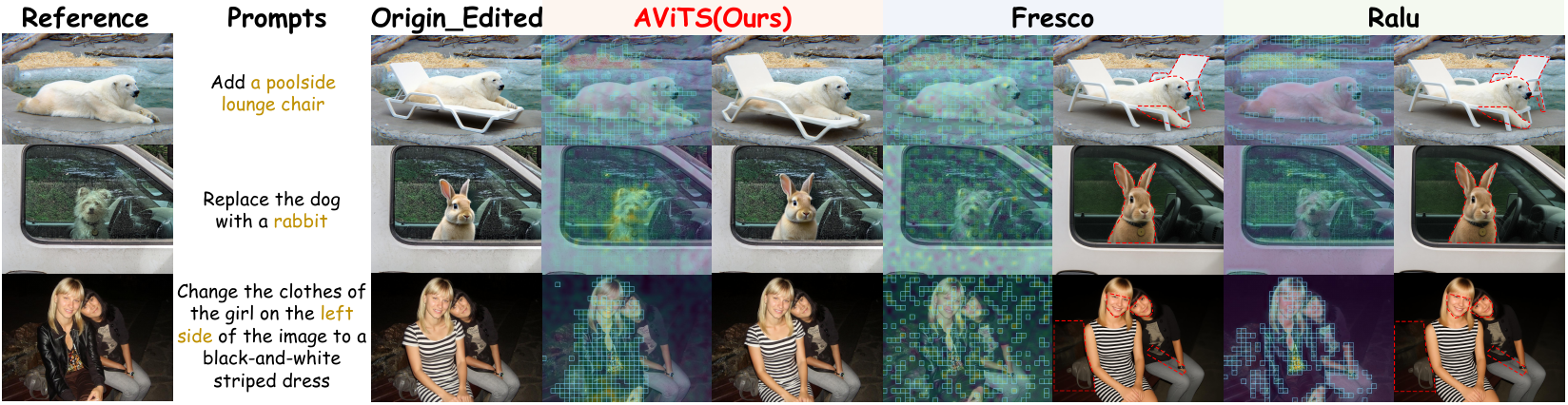}
    \caption{\textbf{Token-importance maps for editing.}
RALU/Fresco tend to allocate importance to irrelevant regions (edges or diffuse background), which leads to \textbf{semantic misalignment} (Row~1), degraded \textbf{color/style consistency} (Row~2), and \textbf{inconsistency in non-edited regions} (Row~3), whereas AViTS concentrates tokens on the instruction-relevant edit areas.}
    \label{fig:compare_hot}
    \vspace{-2mm}
\end{figure}
A central question in dynamic-resolution sampling is which tokens should be prioritized for upsampling under a limited budget.
Existing partial upsampling methods typically assign priorities based on internal latent cues, such as spatial structure (e.g., edge detection on a single decoded frame) or per-channel or single-step feature statistics, without explicitly modeling text-conditioned semantic alignment and cross-timestep representation dynamics.

Figure~\ref{fig:compare_hot} shows token importance heatmaps from our method, RALU~\cite{jeong2025upsample}, and Fresco~\cite{zheng2026sketch} on an editing task.
Ours concentrates on instruction-relevant editing regions; RALU emphasizes edges; Fresco yields more diffuse, less discriminative patterns.
See the caption for a detailed analysis.
These observations motivate us to formalize token importance as a function of both the text condition and the temporal trajectory of latent tokens.

We decompose importance into a spatial component $A_i$ (latent--text alignment,~\cref{sec:spatial}) and a temporal component $V_i$ (cross-step dynamics,~\cref{sec:temporal}):
\begin{equation}
  p_i \;=\; g\!\left(\mathbf{z}_i,\;\mathbf{c},\;
            \bigl\{\mathbf{z}_i^{(t_k)}\bigr\}_{k=1}^{N_T}\right)
  \;=\; g_{\mathrm{sp}}\!\left(\{\alpha_{i,j}^{(h)}\}\right) \;+\; g_{\mathrm{tmp}}\!\left(\bigl\{\mathbf{z}_i^{(t_k)}\bigr\}_{k=1}^{N_T}\right),
  \label{eq:importance_general}
\end{equation}
where $\{\alpha_{i,j}^{(h)}\}$ are the cross-modal attention weights and $\{\mathbf{z}_i^{(t_k)}\}$ are the latent snapshots recorded during the latter steps of Stage~1.
We detail each component.

\subsubsection{Spatial Importance via Latent--Text Alignment}
\label{sec:spatial}

Tokens strongly aligned with the text condition carry conditional semantics and benefit from early high-resolution processing.
We measure this via the cross-modal attention from image tokens to text tokens.

Let $\alpha_{i,j}^{(h)}$ be the attention weight from image token $\mathbf{z}_i$ to text token $\mathbf{c}_j$ in head $h$.
The spatial importance of token $i$ is
\begin{equation}
  A_i \;=\; \frac{1}{|\mathcal{H}|}\sum_{h\,\in\,\mathcal{H}}
             \sum_{j=1}^{L}\alpha_{i,j}^{(h)}.
  \label{eq:spatial}
\end{equation}
The computation of $\{\alpha_{i,j}^{(h)}\}$ depends on whether the backbone uses a multimodal large language model (MLLM) or a DiT with joint image--text blocks.

\paragraph{MLLM-based extraction.}
Models built on MLLMs (e.g., Qwen-Image-Edit) encode image patches and text in a single unified sequence.
The attention matrices at the MLLM layers directly provide an image-to-text submatrix $\boldsymbol{\alpha}^{(h)} \in \mathbb{R}^{M' \times L}$ per head $h$.
We extract this submatrix from a selected layer, aggregate over heads, and sum over the text dimension via Eq.~\eqref{eq:spatial} to obtain $A_i$.

\paragraph{Joint-attention extraction.}
Models without an MLLM (e.g., FLUX-family) use \emph{double-stream} blocks that compute joint self-attention over the concatenated sequence $[\mathbf{C};\mathbf{Z}]$.
The joint query and key are
\begin{equation}
  \mathbf{Q} \;=\; \bigl[\mathbf{Q}_{\mathrm{txt}}\;\|\;\mathbf{Q}_{\mathrm{img}}\bigr],
  \qquad
  \mathbf{K} \;=\; \bigl[\mathbf{K}_{\mathrm{txt}}\;\|\;\mathbf{K}_{\mathrm{img}}\bigr],
  \label{eq:joint_qk}
\end{equation}
where $\mathbf{Q}_{\mathrm{img}},\mathbf{K}_{\mathrm{img}}$ and $\mathbf{Q}_{\mathrm{txt}},\mathbf{K}_{\mathrm{txt}}$ are the image and text projections.
The full attention matrix (before softmax) is $\mathbf{P} = \mathbf{Q}\mathbf{K}^{\top} / \sqrt{d_h}$.
Because fused attention kernels do not expose weights, we register a forward hook on each double block to obtain $\mathbf{Q}$, $\mathbf{K}$, apply the same QK-normalisation as the model, and reconstruct
\begin{equation}
  \boldsymbol{\alpha}^{(h)} \;=\;
  \bigl[\operatorname{softmax}\!\bigl(
        \tilde{\mathbf{Q}}^{(h)}\tilde{\mathbf{K}}^{(h)\top}\big/\sqrt{d_h}\bigr)
  \bigr]_{\mathrm{img}\to\mathrm{txt}},
  \label{eq:dit_attn}
\end{equation}
where $\tilde{\mathbf{Q}}^{(h)}$, $\tilde{\mathbf{K}}^{(h)}$ are $\ell_2$-normalised and the subscript selects rows for image queries and columns for text keys.
We aggregate scores across selected blocks and the $N_T$ collection steps via Eq.~\eqref{eq:spatial}.

\subsubsection{Temporal Importance via Cross-Step Feature Dynamics}
\label{sec:temporal}

The spatial score $A_i$ does not capture how actively a token evolves.
A token whose feature changes substantially across denoising steps is still under construction and benefits from early high-resolution refinement.

We stack the $N_T$ latent snapshots $\{\mathbf{Z}^{(t_k)}\}_{k=1}^{N_T}$ and compute, for each token $i$ and channel $d$, the unbiased sample variance across steps.
Let $\bar{z}_{i,d}$ denote the temporal mean:
\begin{equation}
  \bar{z}_{i,d}
  \;=\; \frac{1}{N_T}\sum_{k=1}^{N_T} z_{i,d}^{(t_k)}.
  \label{eq:zbar}
\end{equation}
The temporal importance of token $i$ is the mean over channels of these variances:
\begin{equation}
  V_i
  \;=\; \frac{1}{D}\sum_{d=1}^{D}
        \frac{1}{N_T - 1}\sum_{k=1}^{N_T}
        \bigl( z_{i,d}^{(t_k)} - \bar{z}_{i,d} \bigr)^{2}.
  \label{eq:step_var}
\end{equation}
Because $V_i$ measures cross-step intra-latent dynamics, it is orthogonal to the cross-modal signal $A_i$.
\subsubsection{Joint Score and Token Selection}
\label{sec:joint}

Both scores are normalised to $[0,1]$ via min--max: $\hat{A}_i = (A_i - \min_i A_i)/(\max_i A_i - \min_i A_i + \varepsilon)$ and analogously $\hat{V}_i$.
The joint importance score is
\begin{equation}
  S_i \;=\; \alpha\,\hat{A}_i \;+\; (1-\alpha)\,\hat{V}_i,
  \qquad \alpha \in [0,1].
  \label{eq:joint}
\end{equation}
We select the top-$K$ tokens:
\begin{equation}
  \mathcal{S} \;=\;
  \operatorname{TopK}\!\bigl(\{S_i\}_{i=1}^{M'},\;
  K{=}\lfloor\rho M'\rfloor\bigr).
  \label{eq:topk}
\end{equation}
A small Gaussian perturbation (on the order of $10^{-6}$) is applied before sorting to break ties.
Orthogonal upsampling and coordinate-bound noise re-injection follow prior practice to preserve cross-stage consistency.

\section{Experiment}
\subsection{Experiment Settings}
\textbf{Model Configurations.}
We conduct experiments on three diffusion-based models: the text-to-image model FLUX.1-dev~\cite{flux2024} and two representative diffusion-based editing models, \textbf{Qwen-Image-Edit}~\cite{wu2025qwen} and \textbf{FLUX.1-Kontext-dev}~\cite{labs2025flux1kontextflowmatching}. 
All experiments are conducted on NVIDIA A800 GPUs for FLUX.1-dev, and on NVIDIA H20 GPUs for Qwen-Image-Edit and FLUX.1-Kontext-dev.

We compare AViTS with a diverse set of acceleration baselines, attention sparsification methods such as SpargeAttention~\cite{zhang2025spargeattn}, feature caching approaches including TeaCache~\cite{liu2025timestep}, token reuse and forecasting methods such as ToCa~\cite{toca}, DuCa~\cite{duca}, FORA~\cite{selvaraju2024fora}, FreqCa~\cite{liu2025freqca}, and TaylorSeer~\cite{liu2025reusing}, as well as spatial resolution scheduling strategies including Bottleneck Sampling~\cite{tian2025training}, RALU~\cite{jeong2025upsample}, and Fresco~\cite{zheng2026sketch}. 
More details are provided in the supplementary materials.

\textbf{Datasets and Metrics.}
Experiments use two representative benchmarks for text-to-image generation and instruction-based image editing. 
For generation evaluation, we adopt the \textbf{DrawBench} benchmark, where the generated samples are assessed using \textbf{ImageReward} and \textbf{CLIP Score} to measure image quality and text–image semantic alignment. 
These metrics jointly evaluate whether the generated images faithfully reflect the input prompt while maintaining high perceptual realism. 
For image editing evaluation, we use the \textbf{GEdit} benchmark, which evaluates instruction-driven editing fidelity and alignment to target modifications under textual and visual guidance. 
Editing quality is measured using \textbf{Semantic Consistency (SC)}, \textbf{Perceptual Quality (PQ)}, and \textbf{Overall Score (OS)}, which together reflect semantic correctness, visual realism, and overall editing performance. 
Computational efficiency is evaluated by the \textbf{number of function evaluations (NFE)}, \textbf{latency}, \textbf{speedup}, and \textbf{FLOPs}.
\begin{table*}[t]
\centering
\caption{\textbf{Quantitative comparison of text-to-image results on FLUX.1-dev.}}
\setlength\tabcolsep{7pt} 
\small
\resizebox{\textwidth}{!}{
\begin{tabular}{l | c | c  c | c  c | c | c }
    \toprule
    \multirow{2}{*}{\centering \bf Method} & \multirow{2}{*}{\centering \bf NFE} & \multicolumn{4}{c|}{\bf Acceleration} & \multirow{2}{*} {\bf Image Reward $\uparrow$} & \multirow{2}{*}{\bf CLIP Score $\uparrow$}\rule{0pt}{2ex} \\
    \cline{3-6}
    & & {\bf Latency(s) $\downarrow$} & {\bf Speed $\uparrow$} & {\bf FLOPs(T) $\downarrow$}  & {\bf Speed $\uparrow$} &  & \\

    \midrule
    {FLUX.1-dev} & 50 & 25.78 & 1.00$\times$ & 3719.50 & 1.00$\times$ & 
    0.9719 \textcolor{gray!70}{\scriptsize (+0.00\%)} & 
    32.325 \textcolor{gray!70}{\scriptsize (+0.00\%)} \\ 
    \midrule

    $60\%$ steps & 30 & 16.63 & 1.55$\times$ & 2231.70 & 1.67$\times$ &
    0.9646 \textcolor{gray!70}{\scriptsize (-0.75\%)} &
    32.232 \textcolor{gray!70}{\scriptsize (-0.28\%)} \\

    SpargeAttention & 50 & 15.44 & 1.67$\times$ & 2150.02 & 1.73$\times$ &
    0.9795 \textcolor{gray!70}{\scriptsize (+0.78\%)} &
    32.312 \textcolor{gray!70}{\scriptsize (-0.04\%)} \\
    
    TeaCache $({l}=0.25)$  & 50 & 14.09 & 1.83$\times$ & 1937.24 & 1.92$\times$ &
    0.9442 \textcolor{gray!70}{\scriptsize (-2.85\%)} &
    32.167 \textcolor{gray!70}{\scriptsize (-0.49\%)} \\

    TaylorSeer $(\mathcal{N}=3)$ & 50 & 9.88 & 2.61$\times$ & 1320.07 & 2.82$\times$ &
    0.9861 \textcolor{gray!70}{\scriptsize (+1.46\%)} &
    32.313 \textcolor{gray!70}{\scriptsize (-0.04\%)} \\

    Bottleneck Sampling & 30 & 11.31 & 2.28$\times$ & 1582.77 & 2.35$\times$ &
    0.9721 \textcolor{gray!70}{\scriptsize (+0.02\%)} &
    32.141 \textcolor{gray!70}{\scriptsize (-0.57\%)} \\

    RALU & 30 & 11.02 & 2.34$\times$ & 1499.79 & 2.48$\times$ &
    0.9626 \textcolor{gray!70}{\scriptsize (-0.96\%)} &
    32.118 \textcolor{gray!70}{\scriptsize (-0.64\%)} \\

    Fresco & 30 & 9.17 & 2.81$\times$ & \textbf{1295.99} & \textbf{2.87$\times$} &
    0.9801 \textcolor{gray!70}{\scriptsize (+0.84\%)} &
    32.125 \textcolor{gray!70}{\scriptsize (-0.62\%)} \\
    
    $\textbf{AViTS}$  &  30 & \bf 8.21 & \textbf{3.14$\times$} & 1401.03  & 2.65$\times$ & 
    \bf 1.0104 \textcolor[HTML]{0F98B0}{\scriptsize \textbf{(+3.96\%)}} &
    \bf 32.476 \textcolor[HTML]{0F98B0}{\scriptsize \textbf{(+0.47\%)}} \\

    \midrule

    $36\%$ steps & 18 & 9.88 & 2.61$\times$ & 1339.02 & 2.77$\times$ &
    0.9553 \textcolor{gray!70}{\scriptsize (-1.71\%)} &
    32.114 \textcolor{gray!70}{\scriptsize (-0.65\%)} \\

    ToCa $(\mathcal{N}=6)$ & 50 & 13.15 & 1.96$\times$ & 924.30 & 4.02$\times$ &
    0.9702 \textcolor{gray!70}{\scriptsize (-0.17\%)} &
    32.083 \textcolor{gray!70}{\scriptsize (-0.75\%)} \\
    
    DuCa $(\mathcal{N}=5)$ & 50 & 8.19 & 3.15$\times$ & 978.76 & 3.80$\times$ &
    0.9855 \textcolor{gray!70}{\scriptsize (+1.40\%)} &
    32.241 \textcolor{gray!70}{\scriptsize (-0.26\%)} \\
    
    TeaCache $({l}=0.8)$  & 50 & 6.63 & 3.89$\times$ & 892.35 & 4.17$\times$ &
    0.8805 \textcolor{gray!70}{\scriptsize (-9.40\%)} &
    31.827 \textcolor{gray!70}{\scriptsize (-1.54\%)} \\

    TaylorSeer $(\mathcal{N}=4)$ & 50 & 9.21 & 2.80$\times$ & 967.91 & 3.84$\times$ &
    0.9857 \textcolor{gray!70}{\scriptsize (+1.42\%)} &
    \bf 32.413 \textcolor{gray!70}{\scriptsize (+0.27\%)} \\

    RALU & 18 & 6.33 & 4.07$\times$ & 904.98 & 4.11$\times$ &
    0.9481 \textcolor{gray!70}{\scriptsize (-2.45\%)} &
    32.074 \textcolor{gray!70}{\scriptsize (-0.78\%)} \\

    Fresco & 18 & 5.72 & 4.51$\times$ & 788.03 & 4.72$\times$ &
    0.9861 \textcolor{gray!70}{\scriptsize (+1.46\%)} &
    31.970 \textcolor{gray!70}{\scriptsize (-1.10\%)} \\
     
    $\textbf{AViTS}$  &  18 & \bf 4.73 & \textbf{5.45$\times$} & \bf774.74 & \textbf{4.80$\times$} &
    \bf 0.9959 \textcolor[HTML]{0F98B0}{\scriptsize \textbf{(+2.47\%)}} &
    32.361 \textcolor[HTML]{0F98B0}{\scriptsize \textbf{(+0.11\%)}} \\

    \midrule

    ToCa $(\mathcal{N}=10)$ & 50 & 7.93 & 3.25$\times$ & 714.66 & 5.20$\times$ &
    0.7055 \textcolor{gray!70}{\scriptsize (-27.41\%)} &
    31.808 \textcolor{gray!70}{\scriptsize (-1.60\%)} \\

    DuCa $(\mathcal{N}=9)$ & 50 & 7.26 & 3.55$\times$ & 690.25 & 5.39$\times$ &
    0.8182 \textcolor{gray!70}{\scriptsize (-15.81\%)} &
    31.759 \textcolor{gray!70}{\scriptsize (-1.75\%)} \\
    
    TeaCache $({l}=1.6)$  & 50 & 3.78 & 6.82$\times$ & 520.54 & 7.15$\times$ &
    0.6423 \textcolor{gray!70}{\scriptsize (-33.91\%)} &
    31.656 \textcolor{gray!70}{\scriptsize (-2.07\%)} \\

    TaylorSeer $(\mathcal{N}=9)$ & 50 & 4.85 & 5.32$\times$ & 596.07 & 6.24$\times$ &
    0.8562 \textcolor{gray!70}{\scriptsize (-11.90\%)} &
    31.653 \textcolor{gray!70}{\scriptsize (-2.08\%)} \\

    FLUX.1-schnell &  8 & 4.21 & 6.12$\times$ & 595.12 & 6.25$\times$ &
    0.9097 \textcolor{gray!70}{\scriptsize (-6.40\%)} &
    \bf 33.837 \textcolor{gray!70}{\scriptsize (+4.68\%)} \\

    RALU & 10 & 3.72 & 6.92$\times$ & 540.62 & 6.88$\times$ &
    0.9289 \textcolor{gray!70}{\scriptsize (-4.42\%)} &
    32.113 \textcolor{gray!70}{\scriptsize (-0.66\%)} \\

    Fresco & 11 & 3.43 & 7.52$\times$ & 486.85 & 7.64$\times$ &
    0.9366 \textcolor{gray!70}{\scriptsize (-3.63\%)} &
    31.897 \textcolor{gray!70}{\scriptsize (-1.32\%)} \\
    
$\textbf{AViTS}$  & 14 &  3.65 &{7.06$\times$} &  586.56 & {6.34$\times$} &
\bf 0.9723 \textcolor[HTML]{0F98B0}{\scriptsize \textbf{(+0.04\%)}} &
32.352 \textcolor[HTML]{0F98B0}{\scriptsize \textbf{(+0.08\%)}} \\

$\textbf{AViTS}$  & 11 &  3.10 &{8.32$\times$} &  486.85 & {7.64$\times$} &
0.9423 \textcolor[HTML]{0F98B0}{\scriptsize \textbf{(-3.05\%)}} &
31.959 \textcolor[HTML]{0F98B0}{\scriptsize \textbf{(-1.13\%)}} \\

$\textbf{AViTS}$  & 9 & \bf 2.64 & \textbf{9.78$\times$} & \bf 380.30 & \textbf{9.78$\times$} &
0.9201 \textcolor[HTML]{0F98B0}{\scriptsize \textbf{(-5.33\%)}} &
32.169 \textcolor[HTML]{0F98B0}{\scriptsize \textbf{(-0.48\%)}} \\

    \bottomrule
\end{tabular}}
\label{table:FLUX}
\footnotesize

\vspace{-10pt}
\end{table*}

\begin{figure}[!b]
    \centering
    \includegraphics[width=\linewidth]{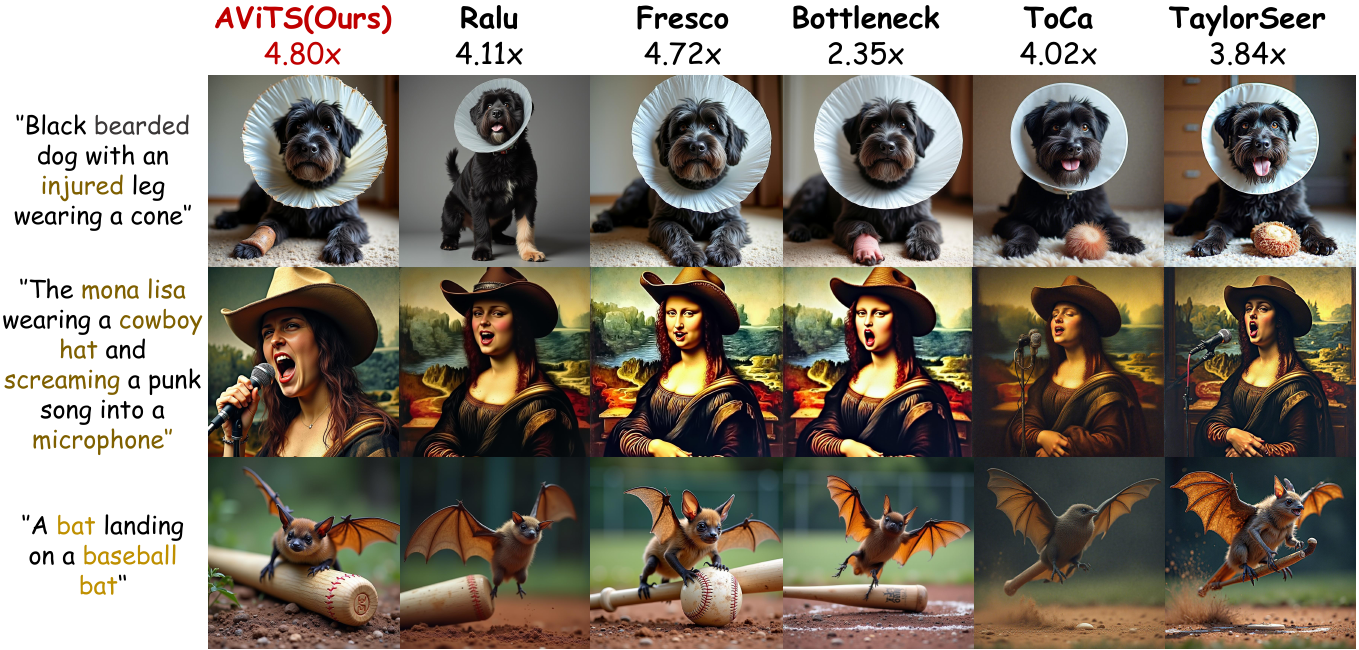}
    \caption{\textbf{Visualization of the image generated by different methods on FLUX.1-dev}. AViTS achieves the best semantic fidelity and the fastest speed (4.80×), surpassing all dynamic-resolution and feature-caching baselines. }
    \label{fig:avits_dynamic2}
\end{figure}

\subsection{Results on Text-to-Image Generation}

As shown in Table~\ref{table:FLUX}, AViTS consistently improves the efficiency--quality trade-off on FLUX.1-dev across a wide range of acceleration regimes.
Under the moderate acceleration setting with NFE=30, AViTS reduces latency to \textbf{8.21}s (\textbf{3.14}$\times$) with \textbf{2.65}$\times$ FLOPs reduction, achieving the lowest latency among competing methods such as Bottleneck Sampling, RALU, and Fresco.
Notably, AViTS also delivers the best generation quality, obtaining the highest ImageReward (\textbf{1.0104}) and CLIP score (\textbf{32.476}), surpassing the original FLUX baseline despite using significantly fewer effective computations.

When the sampling budget is further reduced to NFE=18, AViTS continues to demonstrate strong robustness under more aggressive acceleration, reaching \textbf{4.73}s latency with \textbf{5.45}$\times$ speedup and \textbf{4.80}$\times$ FLOPs reduction while maintaining high visual quality (\textbf{0.9959} ImageReward and \textbf{32.361} CLIP score). 
These results indicate that the proposed token prioritization strategy effectively preserves semantically important structures even when the available computation is significantly reduced. 
In contrast, spatial scheduling methods such as RALU and Fresco show noticeable quality degradation at similar speeds, suggesting that heuristic spatial upsampling strategies struggle to maintain consistent generation fidelity under stronger acceleration. 
AViTS maintains stable generation quality while achieving up to \textbf{9.78}$\times$ latency acceleration, demonstrating a more effective allocation of high-resolution computation to semantically important tokens and enabling a better efficiency–quality trade-off across acceleration regimes (see Fig.~\ref{fig:avits_dynamic2} for qualitative comparisons).

\subsection{Results on Image Editing}
\subsubsection{Results on FLUX.1-Kontext-dev}
\begin{table}[t]
  \centering
  \setlength{\tabcolsep}{4pt}
  \footnotesize
  \caption{Quantitative evaluation results of image editing on the GEdit-bench with \textbf{FLUX.1-Kontext-dev}.}
  \label{tab:flux_kontext_quant}
  \resizebox{\textwidth}{!}{
    \begin{tabular}{l c c c c c c c c}
      \toprule
      \multirow{2}{*}{\textbf{Method}} &
      \multirow{2}{*}{\textbf{NFE}} &
      \multicolumn{4}{c}{\textbf{Acceleration}} &
      \multicolumn{3}{c}{\textbf{GEdit-EN(FULL)}} \\
      \cmidrule(lr){3-6} \cmidrule(lr){7-9}
      & & \textbf{Latency ($\text{s}$) $\downarrow$} & \textbf{Speed $\uparrow$} &
      \textbf{FLOPs ($\text{T}$) $\downarrow$} & \textbf{Speed $\uparrow$} &
      \textbf{SC $\uparrow$} & \textbf{PQ $\uparrow$} & \textbf{OS $\uparrow$} \\
      \midrule

      $100\%$ steps & 50 & 50.20 & 1.00$\times$ & 8299.54 & 1.00$\times$ & \textbf{6.80} & 7.26 & \textbf{6.51} \\
      $60\%$ steps  & 30 & 32.23 & 1.56$\times$ & 4979.72 & 1.67$\times$ & 6.54 & \textbf{7.28} & 6.25 \\
      $20\%$ steps  & 10 & \textbf{10.47} & \textbf{4.79}$\times$ & \textbf{1659.91} & \textbf{5.00}$\times$ & 6.60 & 7.18 & 6.28 \\

      \midrule

      SpargeAttention & 50 & 46.05 & 1.09$\times$ & 5603.68 & 1.48$\times$ & 6.46 & \textbf{7.25} & 6.21 \\
      RALU            & 30 & 23.34  & 2.15$\times$ & 2730.53 & 3.04$\times$ & 7.06 & 7.20 & 6.70 \\
      Bottleneck      & 30 & 18.25  & 2.75$\times$ & 2727.10 & 3.04$\times$ & 6.46 & 6.63 & 6.08 \\
      Fresco          & 50 & 20.48  & 2.45$\times$ & 2361.22 & 3.51$\times$ & 7.02 & 7.12 & 6.65 \\
      \rowcolor{gray!20}
      \textbf{AViTS}  & 30 & \textbf{15.63} & \textbf{3.21}$\times$ & \textbf{2139.98} & \textbf{3.88}$\times$ & \textbf{7.08} & 7.12 & \textbf{6.71} \\

      \midrule

      RALU                 & 18 & 15.30 & 3.28$\times$ & 2123.74 & 3.91$\times$ & 7.05 & 7.17 & 6.69 \\
      Bottleneck           & 18 & 15.21 & 3.30$\times$ & 2274.58 & 3.65$\times$ & 6.86 & 6.77 & 6.43 \\
      ToCa ($\mathcal{N}=8$)        & 50 & 29.56 & 1.70$\times$ & 1841.35 & 4.51$\times$ & 6.43 & 7.25 & 6.12 \\
      TaylorSeer ($\mathcal{N}=6$)  & 50 & 13.95 & 3.60$\times$ & 1660.95 & 5.00$\times$ & 6.47 & \textbf{7.29} & 6.17 \\
      Fresco               & 50 & 17.73 & 2.83$\times$ & 1878.83 & 4.42$\times$ & 7.01 & 7.15 & 6.66 \\
      \rowcolor{gray!20}
      \textbf{AViTS}       & 18 & \textbf{11.51} & \textbf{4.36}$\times$ & \textbf{1544.45} & \textbf{5.37}$\times$ & \textbf{7.07} & 7.12 & \textbf{6.70} \\

      \midrule

      ToCa ($\mathcal{N}=12$)       & 50 & 20.72 & 2.42$\times$ & 1359.61 & 6.10$\times$ & 6.39 & 6.91 & 6.04 \\
      TaylorSeer ($\mathcal{N}=9$)  & 50 & 12.05 & 4.17$\times$ & 1329.02 & 6.24$\times$ & 6.40 & \textbf{6.99} & 6.07 \\
      \rowcolor{gray!20}
      \textbf{AViTS}       & 11 & \textbf{7.25} & \textbf{6.92}$\times$ & \textbf{937.92} & \textbf{8.85}$\times$ & \textbf{7.10} & 6.98 & \textbf{6.57} \\

      \bottomrule
    \end{tabular}
  }
\end{table}

\paragraph{Moderate acceleration regime.}
As shown in Table~\ref{tab:flux_kontext_quant}, in this regime around $\mathbf{3\times}$, AViTS achieves a strong balance between efficiency and editing quality. 
Compared with RALU which reaches $\mathbf{2.15\times}$ speedup, AViTS improves the acceleration to $\mathbf{3.21\times}$ while maintaining better editing performance. 
On GEdit-EN, AViTS obtains the highest semantic consistency with SC \textbf{7.08} and the best overall score of \textbf{6.71}. 
These results indicate that AViTS improves efficiency while preserving reliable semantic alignment.

\paragraph{High and extreme acceleration regimes.}
Under higher acceleration around $\mathbf{4\times}$, AViTS continues to maintain stable editing quality compared with existing methods.
TaylorSeer reaches $\mathbf{3.60\times}$ speedup but its overall score decreases to 6.17.
In contrast, AViTS achieves $\mathbf{4.36\times}$ speedup with stronger editing performance (OS \textbf{6.70}), indicating better allocation of high-resolution computation to important regions.
This suggests importance-aware refinement preserves critical structures even under limited computation.

When the acceleration further increases beyond $\mathbf{5\times}$, AViTS achieves $\mathbf{6.92\times}$ speedup and $\mathbf{8.85\times}$ FLOPs reduction while maintaining strong editing quality with OS \textbf{6.57}.
Despite the substantial reduction in computation, the editing results remain stable across different prompts and image contents.
This observation indicates that AViTS effectively preserves key semantic structures during the denoising process while avoiding unnecessary high-resolution computation.

\begin{table}[t]
\centering
\setlength{\tabcolsep}{4pt}
\small
\caption{Quantitative evaluation results of image editing on the GEdit-bench with \textbf{Qwen-Image-Edit}.}
\label{tab:qwen_edit_quant}
\resizebox{\textwidth}{!}{
\begin{tabular}{l c c c c c c c c c c c}

\toprule
\multirow{2}{*}{\textbf{Method}}
& \multirow{2}{*}{\textbf{NFE}}
& \multicolumn{4}{c}{\textbf{Acceleration}}
& \multicolumn{3}{c}{\textbf{GEdit-CN(FULL)}}
& \multicolumn{3}{c}{\textbf{GEdit-EN(FULL)}} \\

\cmidrule(lr){3-6} \cmidrule(lr){7-9} \cmidrule(lr){10-12}

&
& \textbf{Latency ($s$) $\downarrow$}
& \textbf{Speed $\uparrow$}
& \textbf{FLOPs (T) $\downarrow$}
& \textbf{Speed $\uparrow$}
& \textbf{SC $\uparrow$}
& \textbf{PQ $\uparrow$}
& \textbf{OS $\uparrow$}
& \textbf{SC $\uparrow$}
& \textbf{PQ $\uparrow$}
& \textbf{OS $\uparrow$} \\

\midrule

\textbf{$100\%$ steps}
& 50 & 284.51 & 1.00$\times$ & 28219.71 & 1.00$\times$ & 7.68 & 7.51 & 7.41 & \textbf{7.82} & \textbf{7.54} & \textbf{7.54} \\

\textbf{$60\%$ steps}
& 30 & 172.43 & 1.65$\times$ & 16931.83 & 1.67$\times$ & \textbf{7.70} & \textbf{7.53} & \textbf{7.44} & 7.77 & 7.52 & 7.47 \\

\textbf{$20\%$ steps}
& \textbf{10} & \textbf{58.66} & \textbf{4.85$\times$} & \textbf{5638.18} & \textbf{5.01$\times$} & 7.65 & 7.42 & 7.35 & 7.73 & 7.46 & 7.44 \\

\midrule

SpargeAttention
& 50 & 231.30 & 1.23$\times$ & 16846.78 & 1.67$\times$ & 7.87 & 7.57 & 7.56 & 7.81 & 7.53 & 7.50 \\

Bottleneck
& 30 & 109.38 & 2.60$\times$ & 9954.36 & 2.83$\times$ & 7.44 & \textbf{7.59} & 7.28 & 7.62 & 7.40 & 7.24 \\

RALU
& 30 & 105.87 & 2.69$\times$ & 9286.52 & 3.04$\times$ & 7.83 & 7.58 & 7.55 & 7.83 & 7.52 & 7.52 \\

ToCa ($\mathcal{N}=6$)
& 50 & 172.43 & 1.65$\times$ & 7850.13 & 3.59$\times$ & \textbf{7.89} & 7.50 & \textbf{7.57} & 7.89 & 7.46 & 7.54 \\

\rowcolor{gray!20}
\textbf{AViTS}
& 30 & \textbf{87.54} & \textbf{3.25$\times$} & \textbf{7581.38} & \textbf{3.72$\times$} & \textbf{7.89} & 7.54 & \textbf{7.57} & \textbf{7.95} & \textbf{7.54} & \textbf{7.62} \\

\midrule

RALU
& 18 & 66.94 & 4.25$\times$ & 6200.73 & 4.55$\times$ & 7.89 & \textbf{7.56} & \textbf{7.60} & 7.82 & \textbf{7.52} & 7.51 \\

Bottleneck
& 18 & 85.95 & 3.31$\times$ & 8090.07 & 3.48$\times$ & 7.75 & 7.52 & 7.45 & 7.70 & 7.46 & 7.39 \\

FORA ($\mathcal{N}=5$)
& 50 & \textbf{63.15} & \textbf{4.51$\times$} & 5643.13 & 5.00$\times$ & 7.60 & 7.31 & 7.25 & 7.62 & 7.34 & 7.28 \\

DuCa ($\mathcal{N}=7$)
& 50 & 69.54 & 4.09$\times$ & 5699.89 & 4.95$\times$ & 7.73 & 7.44 & 7.44 & 7.80 & 7.40 & 7.45 \\

TaylorSeer ($\mathcal{N}=6$)
& 50 & 65.66 & 4.33$\times$ & 5643.13 & 5.00$\times$ & 7.53 & 7.40 & 7.25 & 7.60 & 7.37 & 7.30 \\

\rowcolor{gray!20}
\textbf{AViTS}
& \textbf{18} & 63.36 & 4.49$\times$ & \textbf{5382.48} & \textbf{5.24$\times$} & \textbf{7.91} & 7.53 & 7.58 & \textbf{7.87} & \textbf{7.52} & \textbf{7.55} \\

\midrule

FORA ($\mathcal{N}=7$)
& 50 & 52.20 & 5.45$\times$ & 4515.74 & 6.25$\times$ & 7.42 & 7.13 & 7.06 & 7.43 & 7.19 & 7.06 \\

FreqCa ($\mathcal{N}=9$)
& 50 & 51.09 & 5.57$\times$ & 4514.48 & 6.25$\times$ & 7.62 & 7.18 & 7.27 & 7.66 & 7.12 & 7.21 \\

TaylorSeer ($\mathcal{N}=9$)
& 50 & 53.92 & 5.28$\times$ & 4515.74 & 6.25$\times$ & 6.61 & 6.65 & 6.31 & 6.67 & 6.63 & 6.31 \\

\rowcolor{gray!20}
\textbf{AViTS}
& \textbf{11} & \textbf{40.93} & \textbf{6.95$\times$} & \textbf{3262.43} & \textbf{8.65$\times$} & \textbf{7.90} & \textbf{7.47} & \textbf{7.57} & \textbf{7.80} & \textbf{7.48} & \textbf{7.48} \\

\bottomrule
\end{tabular}
}
\end{table}

\begin{figure}[!b]
    \centering
    \includegraphics[width=1\linewidth]{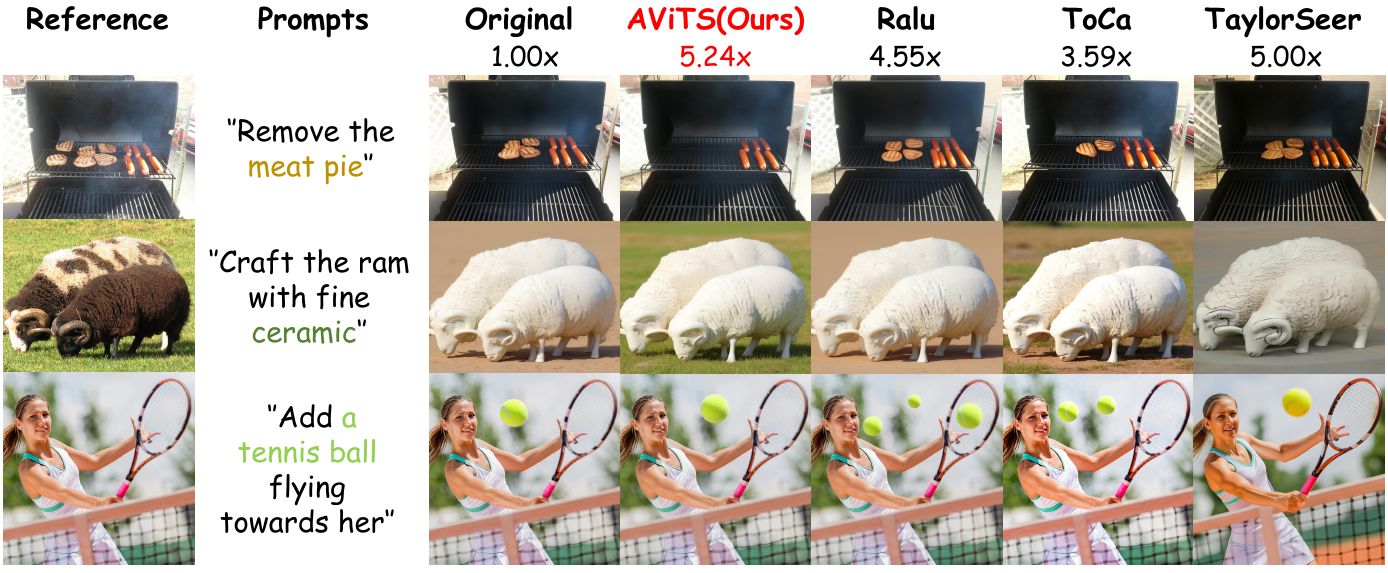}
     \caption{\textbf{Qualitative comparison on GEdit-Bench with Qwen-Image-Edit.} AViTS achieves superior semantic preservation and visual quality at a higher acceleration ratio (\textbf{5.24$\times$}) compared to state-of-the-art baselines.}
    \label{fig:avits_gedit_compare}
\end{figure}

\subsubsection{Results on Qwen-Image-Edit}

\paragraph{Moderate acceleration regime.}
As shown in Table~\ref{tab:qwen_edit_quant}, around $3\times$, AViTS balances efficiency and editing quality. 
Compared with RALU which reaches $2.69\times$ speedup, AViTS increases the acceleration to \textbf{$3.25\times$} while maintaining better editing performance. 
On GEdit-EN, AViTS improves the overall score from 7.52 to \textbf{7.62} and achieves the highest semantic consistency with SC \textbf{7.95}. 
These results indicate that AViTS can further reduce computation while preserving semantic fidelity and perceptual quality (see Fig.~\ref{fig:avits_gedit_compare}).
\paragraph{High and extreme acceleration regimes.}
Under higher acceleration around $4$ to $5\times$, AViTS continues to maintain stable editing quality compared with existing acceleration methods. 
FORA achieves a similar speedup of $4.51\times$ but its editing performance drops to OS 7.25 on GEdit-CN and 7.28 on GEdit-EN. 
In contrast, AViTS achieves $4.49\times$ speedup while maintaining stronger semantic alignment with OS \textbf{7.58} and \textbf{7.55}. 
When the acceleration further increases beyond $5\times$, methods such as FreqCa reach $5.57\times$ speedup but suffer from clear quality degradation. 
AViTS achieves \textbf{$6.95\times$} speedup and \textbf{$8.65\times$} FLOPs reduction while maintaining strong editing performance with OS \textbf{7.57} and \textbf{7.48}. 
These results demonstrate that AViTS scales robustly under aggressive acceleration while preserving stable editing quality.

\subsection{Compatibility with Acceleration Methods}

Table~\ref{table:distill} together with Figure~\ref{fig:accelerate} shows that AViTS is orthogonal to other acceleration axes and can be combined with different compression strategies. 
Without additional training, AViTS combined with feature caching achieves up to \textbf{9}$\times$ acceleration at reduced NFE while maintaining coherent and high-fidelity generations. 
AViTS also remains effective on compressed trajectories. When paired with step distillation, it reaches up to \textbf{14.76}$\times$ acceleration while preserving plausible structures and key visual details. 
Moreover, AViTS composes naturally with quantization such as FLUX.1-dev-int8, providing around \textbf{9}$\times$ speedup while maintaining or slightly improving perceptual quality measured by \textbf{CLIP-IQA}. 
Overall, these results demonstrate that AViTS serves as a plug-and-play token prioritization module that consistently improves the efficiency–quality trade-off when combined with caching, distillation, and quantization.

\begin{table*}[t]
\centering
\caption{\textbf{Quantitative comparison of text-to-image generation} with other accleration methods.}
\setlength\tabcolsep{7pt} 
\small
\resizebox{\textwidth}{!}{
\begin{tabular}{l | c | c  c | c  c | c | c }
    \toprule
    \multirow{2}{*}{\centering \bf Method} & \multirow{2}{*}{\centering \bf NFE} &
    \multicolumn{4}{c|}{\bf Acceleration} &
    \multirow{2}{*} {\bf CLIP-IQA $\uparrow$} &
    \multirow{2}{*} {\bf CLIP Score $\uparrow$}\rule{0pt}{2ex} \\
    \cline{3-6}
    & & {\bf Latency(s) $\downarrow$} & {\bf Speed $\uparrow$} &
    {\bf FLOPs(T) $\downarrow$}  & {\bf Speed $\uparrow$} &  & \\
    \midrule

    {FLUX.1-dev}~\cite{flux2024} & 50 & 25.78 & 1.00$\times$ & 3719.50 & 1.00$\times$ &
    \bf 0.8494 \textcolor{gray!70}{\scriptsize (+0.00\%)} &
    \bf 33.402 \textcolor{gray!70}{\scriptsize (+0.00\%)} \\

    {FreqCa(N=4)}~\cite{liu2025freqca} & 50 & 8.38 & $3.08\times$ & 1116.32 & $3.33\times$ &
    0.8472 \textcolor{gray!70}{\scriptsize (-0.26\%)} &
    32.255 \textcolor{gray!70}{\scriptsize (-3.43\%)} \\

    \rowcolor{gray!20}
    \textbf{AViTS}+Feature Caching  & 18 & \bf 3.02  & \bf 8.54$\times$ & \bf 412.26  & \bf 9.02$\times$ &
     0.8217 \textcolor{gray!30!black}{\scriptsize \textbf{(-3.26\%)}} &
    32.47 \textcolor{gray!30!black}{\scriptsize \textbf{(-2.79\%)}} \\

    \midrule
    {FLUX.1-lite-8B} \cite{flux1-lite} & 28 & 9.16 & 2.81$\times$ & 1291.49 & 2.88$\times$ &
    \bf 0.8617 \textcolor{gray!70}{\scriptsize (+1.45\%)} &
    \bf 32.52 \textcolor{gray!70}{\scriptsize (-2.64\%)} \\

    \rowcolor{gray!20}
    \textbf{AViTS}+Model Distillation  & 14 & \bf 3.09  & \bf 8.34$\times$ & \bf 438.76  & \bf 8.48$\times$ &
    0.8484 \textcolor{gray!30!black}{\scriptsize \textbf{(-0.12\%)}} &
    31.87 \textcolor{gray!30!black}{\scriptsize \textbf{(-4.59\%)}} \\

    \midrule
    {FLUX.1-schnell}~\cite{flux1_schnell} & 8 & 4.20 & 6.14$\times$ & 595.12 & 6.25$\times$ &
    0.8587 \textcolor{gray!70}{\scriptsize (+1.09\%)} &
    33.69 \textcolor{gray!70}{\scriptsize (+0.86\%)} \\

    {FLUX.1-schnell}~\cite{flux1_schnell} & 6 & 3.49 & 7.39$\times$ & 439.34 & 8.47$\times$ &
    0.8703 \textcolor{gray!70}{\scriptsize (+2.46\%)} &
    \bf 33.75 \textcolor{gray!70}{\scriptsize (+1.04\%)} \\

    \rowcolor{gray!20}
    \textbf{AViTS}+Step Distillation  & 8 & 2.55 & 10.11$\times$ & 362.35 & 10.26$\times$ &
    \bf 0.8784 \textcolor{gray!30!black}{\scriptsize \textbf{(+3.41\%)}} &
    31.78 \textcolor{gray!30!black}{\scriptsize \textbf{(-4.86\%)}} \\

    \rowcolor{gray!20}
    \textbf{AViTS}+Step Distillation  & 6 & \bf 1.76  & \bf 14.65$\times$ & \bf 252.00  & \bf 14.76$\times$ &
    0.8709 \textcolor{gray!30!black}{\scriptsize \textbf{(+2.53\%)}} &
    32.20 \textcolor{gray!30!black}{\scriptsize \textbf{(-3.60\%)}} \\

    \midrule
    {FLUX.1-dev-int8}~\cite{flux1dev_torchao_int8} & 50 & 14.01 & 1.84$\times$ & 1888.07 & 1.97$\times$ &
    0.8498 \textcolor{gray!70}{\scriptsize (+0.05\%)} &
    \bf 33.51 \textcolor{gray!70}{\scriptsize (+0.32\%)} \\

    \rowcolor{gray!20}
    \textbf{AViTS}+Quantization  & 30 & 5.28 & 4.88$\times$ & 752.52 & 4.94$\times$ &
    0.8542 \textcolor{gray!30!black}{\scriptsize \textbf{(+0.57\%)}} &
    32.25 \textcolor{gray!30!black}{\scriptsize \textbf{(-3.45\%)}} \\

    \rowcolor{gray!20}
    \textbf{AViTS}+Quantization  & 18 & \bf 2.90  & \bf 8.89$\times$ & \bf 413.40  & \bf 9.00$\times$ &
    \bf 0.8723 \textcolor{gray!30!black}{\scriptsize \textbf{(+2.70\%)}} &
    32.08 \textcolor{gray!30!black}{\scriptsize \textbf{(-3.96\%)}} \\

    \bottomrule
\end{tabular}
\label{table:distill}}
\end{table*}

\subsection{Ablation Study}
\begin{table*}[!t]
\centering
\caption{\textbf{Quantitative comparison of text-to-image gen.} on FLUX.1-dev.}
\setlength\tabcolsep{7pt}
\small
\resizebox{\textwidth}{!}{
\begin{tabular}{l | c | c  c | c | c }
\toprule
\multirow{2}{*}{\centering \bf Method} &
\multirow{2}{*}{\centering \bf NFE} &
\multicolumn{2}{c|}{\bf Acceleration} &
\multirow{2}{*}{\bf Image Reward $\uparrow$} &
\multirow{2}{*}{\bf CLIP Score $\uparrow$} \rule{0pt}{2ex} \\
\cline{3-4}
& & {\bf Latency(s) $\downarrow$} & {\bf Speed $\uparrow$} & & \\
\midrule
FLUX.1-dev & 50 & 25.78 & 1.00$\times$ &
0.9719 \textcolor{gray!70}{\scriptsize (+0.00\%)} &
32.325 \textcolor{gray!70}{\scriptsize (+0.00\%)} \\
\midrule
Random & 30 & 5.25 & 4.91$\times$ &
0.9257 \textcolor{gray!70}{\scriptsize (-4.75\%)} &
31.488 \textcolor{gray!70}{\scriptsize (-2.59\%)} \\

Evenly & 30 & 5.25 & 4.91$\times$ &
0.9689 \textcolor{gray!70}{\scriptsize (-0.31\%)} &
32.012 \textcolor{gray!70}{\scriptsize (-0.97\%)} \\

Edge detection & 30 & 5.76 & 4.48$\times$ &
0.9512 \textcolor{gray!70}{\scriptsize (-2.13\%)} &
32.074 \textcolor{gray!70}{\scriptsize (-0.78\%)} \\

Fresco & 30 & 5.72 & 4.51$\times$ &
0.9861 \textcolor{gray!70}{\scriptsize (+1.46\%)} &
31.970 \textcolor{gray!70}{\scriptsize (-1.10\%)} \\

AViTS (only attention) & 30 & 4.72 & 5.46$\times$ &
0.9875 \textcolor{gray!70}{\scriptsize (+1.60\%)} &
32.251 \textcolor{gray!70}{\scriptsize (-0.23\%)} \\

AViTS (only step-variance) & 30 & \textbf{4.67} & \textbf{5.52$\times$} &
0.9846 \textcolor{gray!70}{\scriptsize (+1.31\%)} &
32.202 \textcolor{gray!70}{\scriptsize (-0.38\%)} \\

\rowcolor{gray!20}
\textbf{AViTS (mix)} & 30 & 4.73 & 5.45$\times$ &
\textbf{0.9959} \textcolor{gray!70}{\scriptsize (+2.47\%)} &
\textbf{32.361} \textcolor{gray!70}{\scriptsize (+0.11\%)} \\

\midrule
\bottomrule
\end{tabular}}
\label{table:ablation}

\vspace{3mm}
\centering
\includegraphics[width=\linewidth]{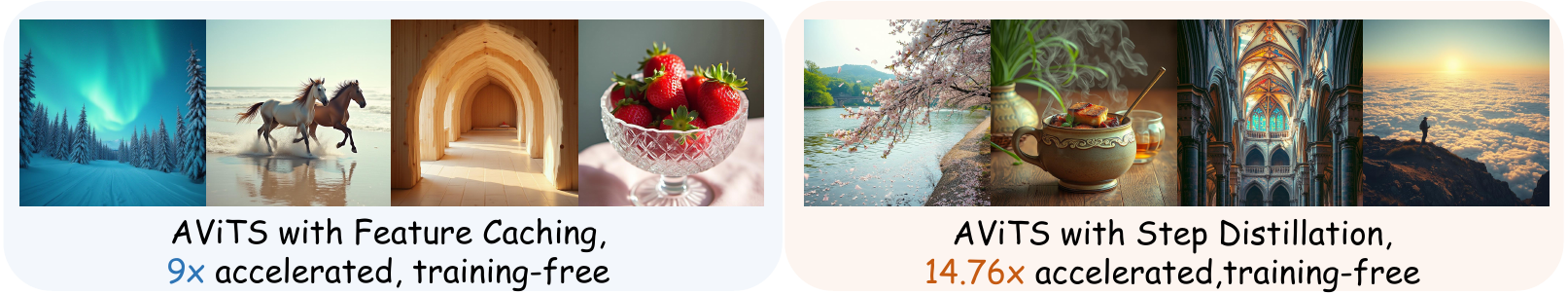}
\captionof{figure}{AViTS composes with feature caching and step distillation for further speedup.}
\label{fig:accelerate}
\end{table*}
We conduct an ablation on FLUX.1-dev under the \emph{same} compute budget (\( \mathrm{NFE}=30 \)) and the \emph{same} upsampling ratio \( \rho \) (Table~\ref{table:ablation}), isolating the effect of \emph{which} tokens are prioritized. 
The training-free allocations, including \textit{Random} and \textit{Evenly}, achieve similar speedups (\(\sim\)4.9\(\times\)) but cause clear drops in ImageReward and CLIP Score. 
Low-level heuristics such as \textit{Edge detection} and \textit{Fresco} remain suboptimal under the same \( \rho \), showing limited alignment with text semantics and reduced CLIP Score. 
In contrast, our proposed importance cues are both effective. Using \textbf{Attention} or \textbf{Step-variance} alone already reaches \(\sim\)5.5\(\times\) acceleration with better quality retention, where step-variance is slightly faster while attention better preserves conditional semantics. 
Finally, \textbf{AViTS (mix)} achieves the best trade-off at fixed \( \rho \), reaching 5.45\(\times\) speedup with the highest ImageReward (0.9959) and CLIP Score (32.361), validating the complementarity between semantic alignment and cross-step dynamics.

\section{Conclusion}
We propose \textbf{AViTS}, a spatiotemporal token selection framework for dynamic-resolution DiTs. AViTS fuses \emph{latent--text attention} (spatial importance) and \emph{cross-step feature variation} (temporal importance) to prioritize critical tokens for early refinement, reducing redundant high-resolution computation while preserving quality. Experiments on DrawBench and GEdit show strong speedups with stable fidelity on FLUX.1-dev, FLUX.1-Kontext-dev, and Qwen-Image-Edit, and AViTS composes well with distillation, quantization, and feature caching.

\section*{Acknowledgements}

This paper was partially sponsored by Terminal Intelligent Computing Division - Alibaba Cloud.
%
%
\bibliographystyle{splncs04}
\bibliography{main}

\clearpage
\appendix
\section{Additional Ablation Studies}
\begin{figure}[htbp]
    \centering
    \includegraphics[width=\linewidth]{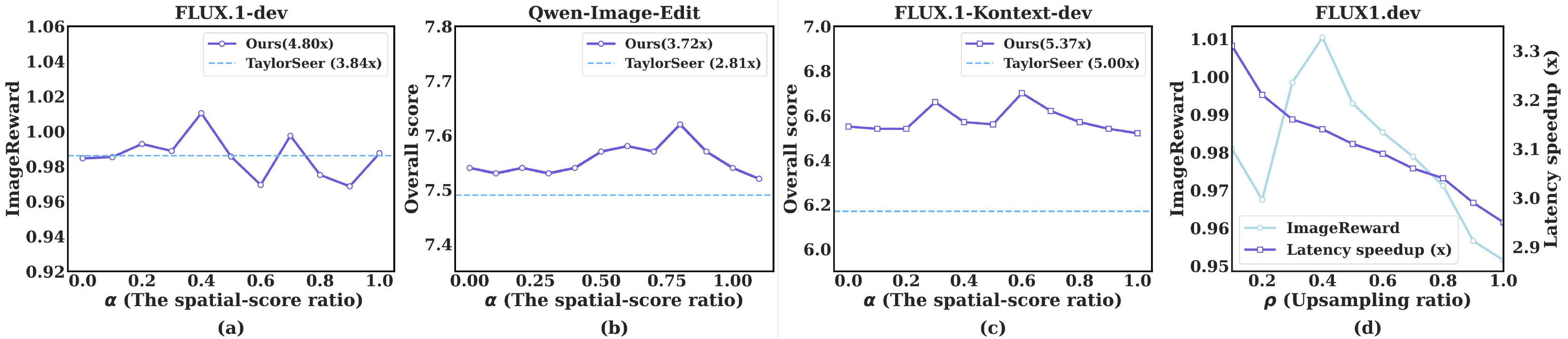}
    \caption{\textbf{Ablation on fusion ratio and upsampling ratio.}
\textbf{(a--c)} Impact of the fusion weight $\alpha$ (spatial attention vs.\ temporal step-variance) on three models: FLUX.1-dev (ImageReward), Qwen-Image-Edit (Overall score), and FLUX.1-Kontext-dev (Overall score). $\alpha{=}0$ uses only step-variance, $\alpha{=}1$ uses only attention. AViTS remains strong across a wide range of $\alpha$ and outperforms TaylorSeer at comparable speedups (shown in legends), indicating complementarity between semantic alignment and cross-step dynamics.
\textbf{(d)} Speed--quality trade-off on FLUX.1-dev by varying the upsampling ratio $\rho$: increasing $\rho$ improves ImageReward but reduces latency speedup, revealing a clear Pareto frontier for selecting the high-resolution budget.}
    \label{fig:xiaorong}
\end{figure}

\begin{figure}[t]
    \centering
    \includegraphics[width=\linewidth]{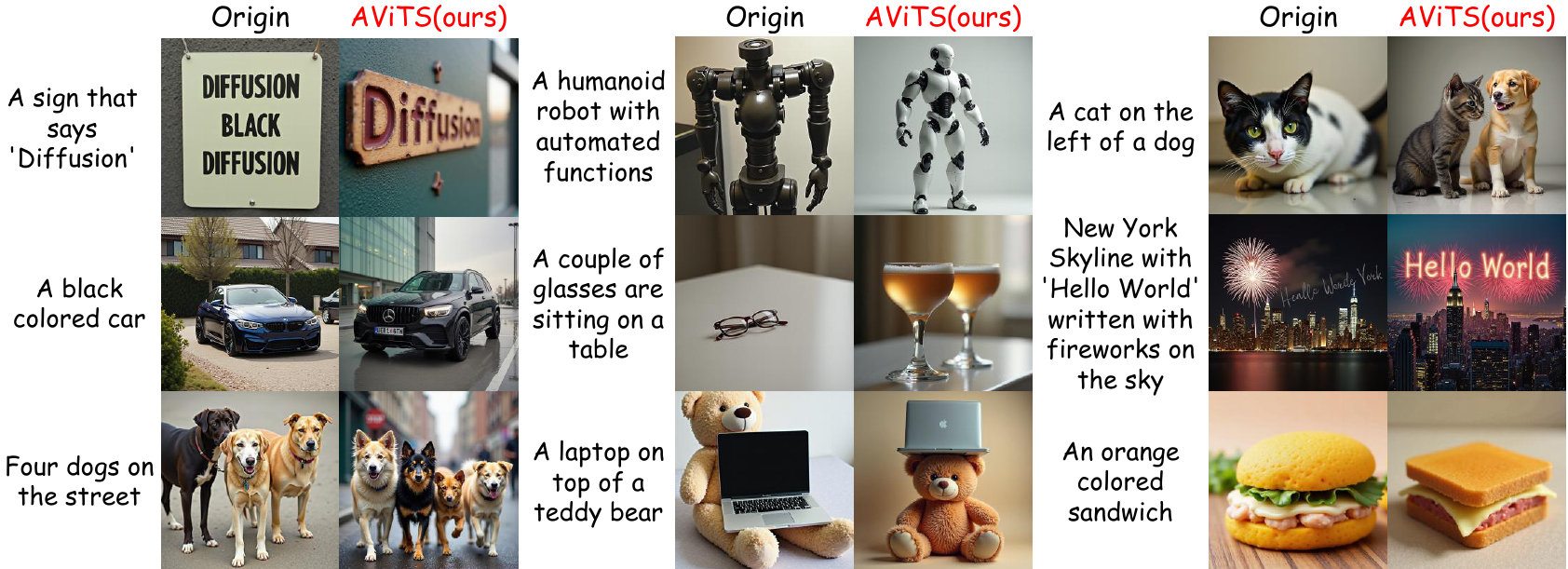}
   \caption{\textbf{Qualitative comparison for high-resolution generation (2048$\times$2048).}
Compared with the original FLUX sampling at the target resolution, AViTS produces more faithful and coherent results, with better color consistency and fewer artifacts, especially for prompts involving text rendering and multi-object relations.}
    \label{fig:avits_2048}
\end{figure}

\paragraph{Effect of fusion ratio $\alpha$ and upsampling ratio $\rho$.}
We study two key hyper-parameters of AViTS: the fusion weight $\alpha$ between spatial importance (latent--text attention) and temporal importance (step-variance), and the upsampling ratio $\rho$ controlling the high-resolution token budget. Figure~\ref{fig:xiaorong}(a--c) shows that AViTS performs well across a broad range of $\alpha$ on all three models, suggesting that the two cues are complementary rather than interchangeable. Intermediate values of $\alpha$ typically yield the best or near-best scores, indicating that relying on either attention-only ($\alpha{=}1$) or step-variance-only ($\alpha{=}0$) is suboptimal. Importantly, AViTS consistently matches or surpasses TaylorSeer under comparable speedups (see legends), showing that importance-aware selective upsampling provides a more reliable efficiency--quality trade-off across backbones. Figure~\ref{fig:xiaorong}(d) further reveals the expected monotonic trade-off with $\rho$: allocating more tokens to early high-resolution refinement improves ImageReward but reduces latency speedup. We thus choose $\rho$ in the middle range to sit on the Pareto frontier.
\paragraph{High-resolution generation.}
Scaling Diffusion Transformers to high output resolutions can be challenging: directly denoising at the target resolution may reduce coherence and yield less stable cues (e.g., region-wise texture or color inconsistency). Table~\ref{tab:resolution_compare} reflects this trend on FLUX: as the resolution increases from 1024$\times$1024 to 1440$\times$1440 and 2048$\times$2048, the baseline quality drops sharply, while AViTS consistently achieves higher ImageReward (0.9008 vs.\ 0.7385 at 1440$\times$1440; 0.8905 vs.\ 0.8371 at 2048$\times$2048) with \(\sim\)5.2\(\times\)–5.5\(\times\) speedup. The qualitative results in Fig.~\ref{fig:avits_2048} support the same conclusion: compared with direct high-resolution sampling, AViTS better preserves prompt-critical semantics (attributes and relations) and produces more globally consistent layouts, with fewer localized artifacts and color drifts. We attribute these improvements to allocating early high-resolution computation only to the most important tokens, which reduces refinement of low-importance regions and stabilizes the trajectory. Nevertheless, extremely high-resolution generation is still challenging for current DiTs; AViTS mainly \emph{reduces} the frequency and severity of such failures rather than completely eliminating them, especially for prompts requiring dense fine-grained details or precise text rendering.
\begin{table}[!ht]
\centering
\caption{\textbf{Comparison at different resolutions.}}
\label{tab:resolution_compare}
\vspace{1mm}

\small
\setlength{\tabcolsep}{6pt}
\renewcommand{\arraystretch}{1.1}

\resizebox{\linewidth}{!}{%
\begin{tabular}{c|c|c|c|c}
\hline
\textbf{Resolution} & \textbf{Method} & \textbf{Latency(s) $\downarrow$} & \textbf{Speed $\uparrow$} & \textbf{ImageReward $\uparrow$} \\
\hline
\multirow{2}{*}{\textbf{1024$\times$1024}}
& FLUX  & 25.78 & 1.00$\times$ & 0.9719 \\
& AViTS & \textbf{4.73} & \textbf{5.45$\times$} & \textbf{0.9959} \\
\hline
\multirow{2}{*}{\textbf{1440$\times$1440}}
& FLUX  & 52.32 & 1.00$\times$ & 0.7385 \\
& AViTS & \textbf{9.72} & \textbf{5.38$\times$} & \textbf{0.9008} \\
\hline
\multirow{2}{*}{\textbf{2048$\times$2048}}
& FLUX  & 120.03 & 1.00$\times$ & 0.8371 \\
& AViTS & \textbf{22.87} & \textbf{5.25$\times$} & \textbf{0.8905} \\
\hline
\end{tabular}%
}
\end{table}

\section{Supplementary visualizations}
\begin{figure}[t]
    \centering
    \includegraphics[width=\linewidth]{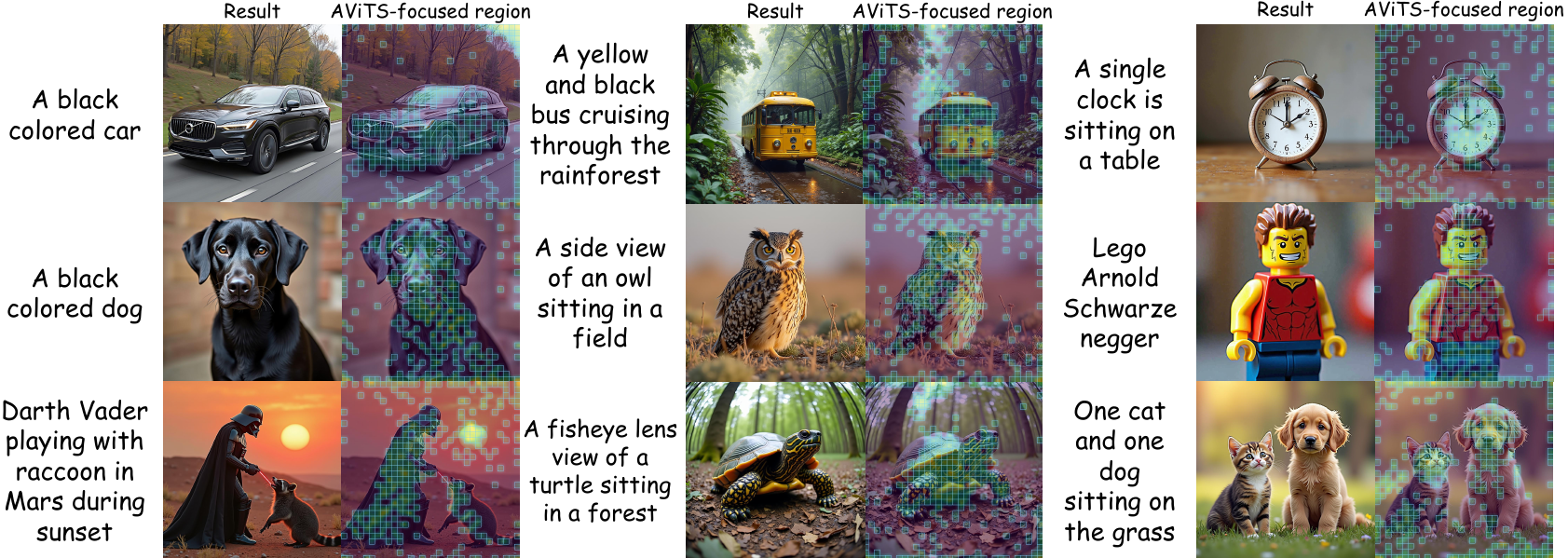}
    \caption{\textbf{Additional visualizations of AViTS token selection for text-to-image generation.}
For each prompt, we show the generated result and the corresponding \textit{AViTS-focused} token regions (highlighted overlays) selected by the final spatiotemporal importance score.
AViTS consistently assigns more high-resolution budget to semantically critical regions (e.g., object parts/attributes, text-bearing areas, and salient regions), showing strong correlation between token prioritization and prompt intent.}

    \label{fig:avits_mix}
\end{figure}

\begin{figure}[htbp]
    \centering
    \includegraphics[width=\linewidth]{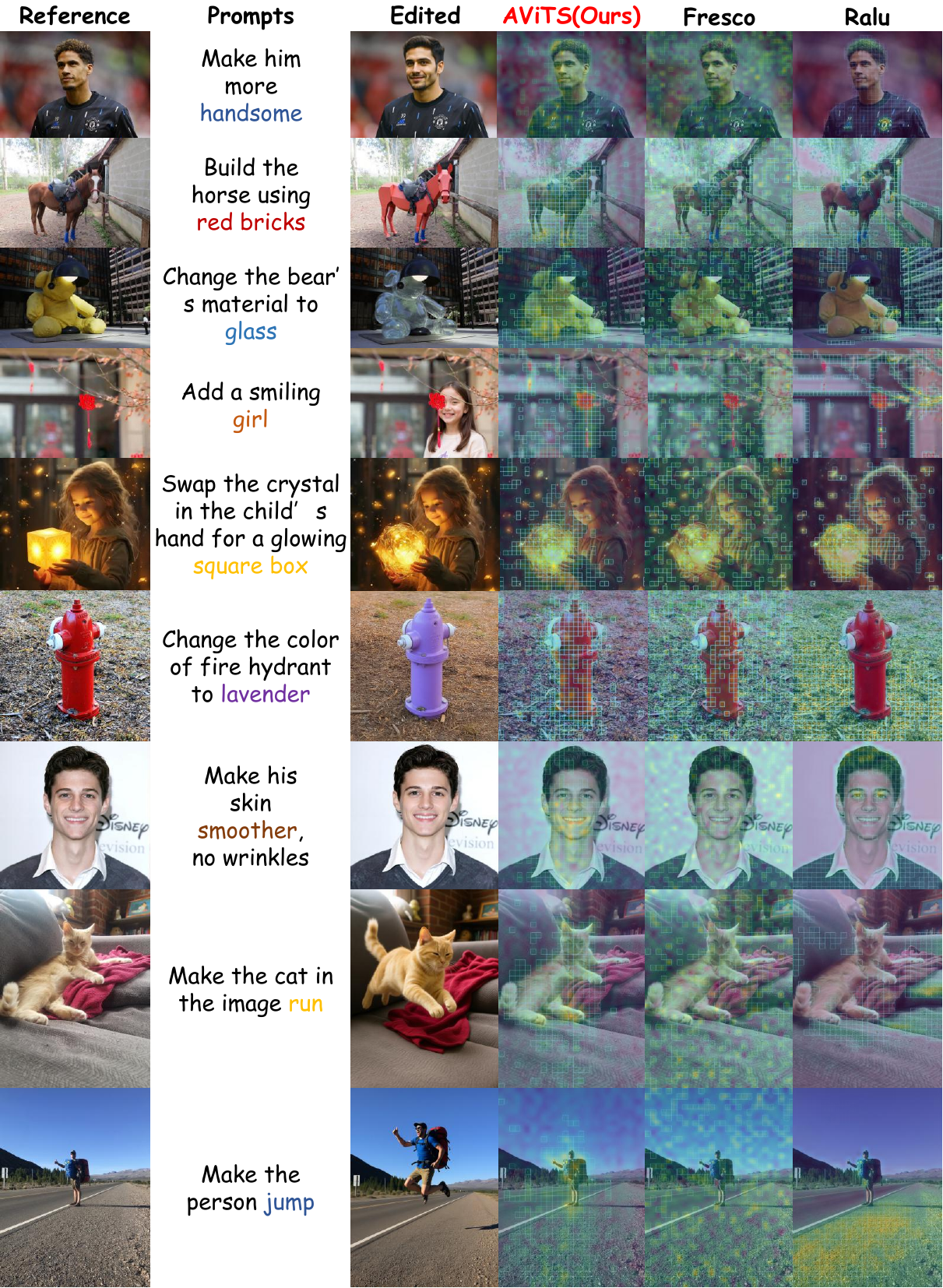}
    \caption{\textbf{Additional editing examples and token-allocation comparison.}
We compare AViTS with Fresco and RALU on diverse editing instructions.
Along with the edited outputs, we visualize the tokens prioritized for high-resolution refinement.
AViTS focuses on instruction-relevant regions (e.g., the target object/attribute to be modified) and assigns fewer tokens to non-edited areas, whereas Fresco/RALU often over-allocate tokens to edges or diffuse background regions, which can lead to weaker edit fidelity or reduced consistency outside the edited region.}

    \label{fig:avits_hot3}
\end{figure}

\paragraph{Overview of supplementary visualizations.}
We provide additional qualitative visualizations to complement the main paper.
Fig.~\ref{fig:avits_mix} shows more examples of the final token subset selected by AViTS for \emph{text-to-image} generation, where the highlighted \textit{AViTS-focused} regions correspond well to the prompt (e.g., object attributes, text rendering, and multi-object relations).
Fig.~\ref{fig:avits_hot3} further presents more \emph{editing} cases, comparing AViTS with prior partial-upsampling heuristics (RALU and Fresco). Consistent with the main text, AViTS concentrates high-resolution computation on instruction-relevant regions while allocating fewer tokens to non-edited areas, leading to better edit fidelity and stronger background/identity preservation.
These figures show that AViTS's token prioritization aligns computation with semantic intent across generation and editing.

\end{document}